\documentclass[a4paper, 10pt, conference]{ieeeconf}      
\usepackage{FG2019}
\usepackage{graphicx,subfigure}
\usepackage{mathtools}
\usepackage{multirow}

\FGfinalcopy 

\IEEEoverridecommandlockouts                              
\def\FGPaperID{72} 

\title{\LARGE \bf
	Bounded Residual Gradient Networks (BReG-Net) for Facial Affect Computing
}

\author{\parbox{16cm}{\centering
    {\large Behzad Hasani, Pooran Singh Negi, and Mohammad H. Mahoor}\\
    {\normalsize
    Department of Electrical \& Computer Engineering, University of Denver, USA\\}}
}

\begin{document}
\IEEEoverridecommandlockouts\pubid{\makebox[\columnwidth]{978-1-7281-0089-0/19/\$31.00~\copyright{}2019 IEEE \hfill}
	\hspace{\columnsep}\makebox[\columnwidth]{ }}	
	\ifFGfinal
	\thispagestyle{empty}
	\pagestyle{empty}
	\else
	\author{Anonymous FG 2019 submission\\ Paper ID \FGPaperID \\}
	\pagestyle{plain}
	\fi
	\maketitle

	\begin{abstract}
		Residual-based neural networks have shown remarkable results in various visual recognition tasks including Facial Expression Recognition (FER). Despite the tremendous efforts have been made to improve the performance of FER systems using DNNs, existing methods are not generalizable enough for practical applications. This paper introduces Bounded Residual Gradient Networks (BReG-Net) for facial expression recognition, in which the shortcut connection between the input and the output of the ResNet module is replaced with a differentiable function with a bounded gradient. This configuration prevents the network from facing the vanishing or exploding gradient problem. We show that utilizing such non-linear units will result in shallower networks with better performance. Further, by using a weighted loss function which gives a higher priority to less represented categories, we can achieve an overall better recognition rate. The results of our experiments show that BReG-Nets outperform state-of-the-art methods on three publicly available facial databases in the wild, on both the categorical and dimensional models of affect.
	\end{abstract}

	\section{INTRODUCTION}

	Facial expressions are one of the most important nonverbal channels for expressing internal emotions during face-to-face communication. Six expressions of anger, disgust, fear, happiness, sadness, and surprise are defined as the basic emotional expressions by Ekman \emph{et al.}~\cite{c5}. Automated Facial Expression Recognition (FER) has been a topic of study for decades. Although there have been many achievements in developing automated FER systems, the majority of existing methods lack the required generalization due to a use of controlled data in developing methods~\cite{c2}. This is predominant because there are significant  variations in facial images owing to variable scene lighting, background variation, camera view, and subjects' head pose, gender, and ethnicity~\cite{c53}. A comprehensive way of studying facial expressions is to approach the task through the concept of \textit{affective computing}. Affect is a psychological term for describing the external exhibition of internal emotions and feelings. Affective computing attempts to develop systems that can interpret and estimate human affects through different channels (e.g. visual, auditory, biological signals, etc.)~\cite{tao2005affective}.

	The dimensional modeling of affect can distinguish between subtle differences in exhibiting affect
	and encode small changes in the intensity of each emotion on a continuous scale, such as \textit{valence}  and \textit{arousal} where valence shows how positive or negative an emotion is, and arousal indicates how much an event is intriguing/agitating or calming/soothing~\cite{russell1980circumplex}. This paper focuses on  developing automated algorithms for computation of the categorical and dimensional models of affect.
\begin{figure}[tb]
	\centering
	\subfigure[ResNet-110]{\label{fig:ResNet-110}\includegraphics[width=\linewidth]{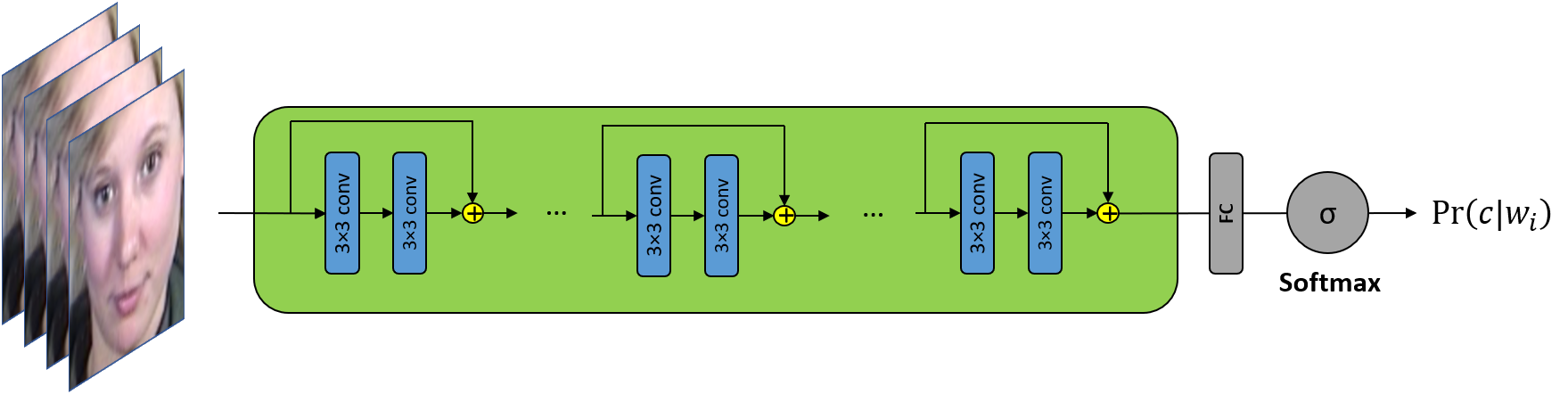}}
	\\
	\subfigure[BReG-Net-39]{\label{fig:CResNet}\includegraphics[width=0.6\linewidth]{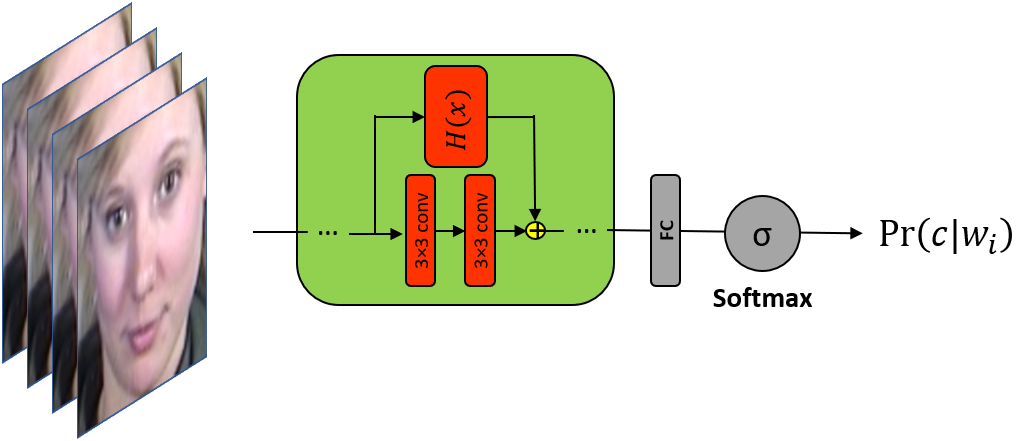}}
	%
	\caption{Comparison of a) ResNet-110 and b) BReG-Net-39. RestNet-110 has more layers, while is slower and less accurate. BReG-Net-39 is shallower, faster, and more accurate. }\label{fig:overview}
\end{figure}

	In the field of machine learning, one of the main tasks is to optimize a function or distribution estimation with respect to a defined measure. Based on the connectionist principle~\cite{rumelhart1986parallel}, deep neural networks allow us to build very complex classes of functions.
	A wide variety of network topologies and activation functions have been proposed in the recent years and they seem to play a crucial role in design and improving the underline class of reproducible functions available to DNNs. To pave the way of training  very deep DNNs, current methods focus on improving neuron saturation or the efficiency of the gradient  flow across various network’s layers. Such approaches are evident in the ReLU class of non-linear functions, and the use of identity mappings in Deep Residual Networks~\cite{he2016identity}. While having deeper architectures has shown to improve the result of recognition, one possibility is to design more complex neurons to extract more useful information at each layer of the network which results in shallower networks and less parameters but more comprehensive information and a higher recognition rate. 
	
	This paper proposes and evaluates BReG-Net (Figure~\ref{fig:overview}), in which the aforementioned identity mapping is replaced with a differentiable function with a bounded gradient that results in a shallower network with a considerably better recognition rate. We evaluate our proposed method using three in the wild facial expression databases (AffectNet~\cite{mollahosseini2017affectnet}, Affect-in-the-wild~\cite{zafeiriou2017aff}, and FER2013~\cite{FER2013}) in computation of both the categorical and dimensional models of affect.
	
	\begin{figure*}[tb]
		\centering
		\subfigure[ResNet]{\label{fig:ResNet}\includegraphics[width=0.25\linewidth]{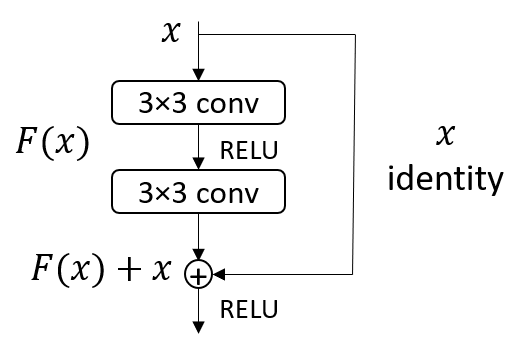}}
		\subfigure[BReG-Net]{\label{fig:CResNet}\includegraphics[width=0.25\linewidth]{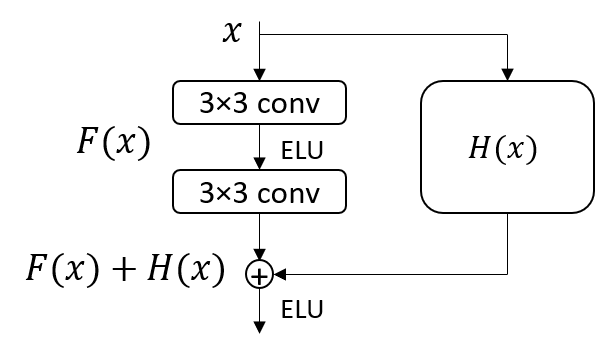}}
		\subfigure[BReG-Net down sampling]{\label{fig:CResNetDS}\includegraphics[width=0.25\linewidth]{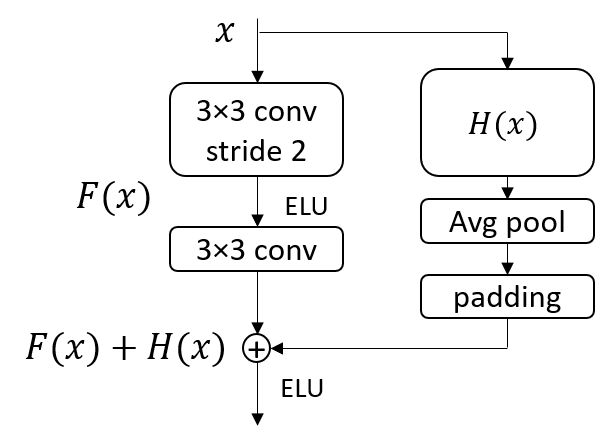}}
		\caption{Block diagram of a) ResNet b) BReG-Net and c) BReG-Net with Down-Sampling building blocks }\label{fig:functionsPlot}
	\end{figure*}

	\section{RELATED WORK}\label{sec:relatedwork}
	\subsection{Facial expression recognition}
	
	In recent years, ``Convolutional Neural Networks" (CNNs) have become the most popular approach in the field of computer vision and pattern recognition. AlexNet and GoogLeNet are among the first successful architectures proposed based on CNNs. AlexNet consists of several convolution layers followed by max-pooling layers and Rectified Linear Units (ReLUs). Szegedy \emph{et al.}~\cite{c18} introduced GoogLeNet which is composed of multiple  ``Inception"  layers. Inception applies several convolutions on the feature map in different scales which extends the model both in depth and width. Mollahosseini \emph{et al.}~\cite{c4,c53} have used the Inception layer for the task of facial expression recognition and achieved state-of-the-art results. Following the success of Inception layers, several variations of them have been proposed~\cite{c20}. Moreover, Inception layer is combined with a residual unit introduced by He \emph{et al.}~\cite{c6} and shows that the resulting architecture accelerates the training of Inception networks significantly~\cite{c23}. Hasani \emph{et al.} proposed a modification of ResNets for the task of facial expression recognition~\cite{hasani2017facial} and valence/arousal prediction of emotions~\cite{hasani2017facial_dimensional}. While these methods use very deep architectures, the question of whether having a more complex building block results in a shallower and more efficient network remains unanswered. In the following, we will review some of the works that have looked into this concept.

	\subsection{Dimensional model of affect}
	A few studies have been conducted on the dimensional model of affect in the literature. Nicolaou \emph{et al.}~\cite{nicolaou2011continuous} trained bidirectional Long Short Term Memory (LSTM) architecture on multiple engineered features extracted from audio, facial geometry, and shoulders. They achieved Root Mean Square Error (RMSE) of 0.15 and Correlation Coefficient (CC) of 0.79 for valence as well as RMSE of 0.21 and CC of  0.64 for arousal. He \emph{et al.}~\cite{he2015multimodal} won the AVEC 2015 challenge by training multiple stacks of bidirectional LSTMs (DBLSTM-RNN) on engineered features extracted from audio (LLDs features), video (LPQ-TOP features), 52 ECG features, and 22 EDA features. They achieved RMSE of 0.104 and CC of 0.616 for valence as well as RMSE of 0.121 and CC of 0.753  for  arousal. Koelstra \emph{et al.}~\cite{koelstra2012deap} trained Gaussian naive Bayes classifiers  on EEG, physiological signals, and multimedia features by binary classification of low/high categories for arousal, valence, and liking on their proposed database DEAP. They achieved F1-score of 0.39, 0.37, and 0.40 on arousal, valence, and liking categories respectively.

	\section{PROPOSED METHOD}\label{sec:prposedmethod}
	
	In this paper, we propose a residual-based network in which the shortcut connection between the input and the output of the module is replaced with a differentiable function with bounded gradient. 
	In the following, we explain each of the aforementioned concepts in detail.
	\subsection{BReG-Net}
	
	The shortcut path in the ResNet module, which connects the input and output of the residual unit proposed, results in accelerating the convergence of the loss and simultaneously prevents the problem of vanishing/exploding gradient. The residual unit can be expressed as:
	\begin{equation}
	\begin{split}
	y_l = H(x_l&) + F (x_l, W_l)\\
	x_{l+1} &= f(y_l)
	\end{split}
	\label{eq:residual_formula}
	\end{equation}
	where $x_l$ and $x_{l+1}$ are the input and the output of the $l$-th unit and $F$ is a residual function. In~\cite{he2015deep}, $H(x_l)=x_l$ is a shortcut path, and $f$ is an ReLU function. Later on in~\cite{he2016identity}, different combination of components both on $F$ and the shortcut was investigated. Hasani \emph{et al.}~\cite{hasani2017facial} proposed a 3D ResNet based model for the task of facial expression recognition in which the shortcut was replaced with element-wise multiplication of the weight function $\omega$ and the input layer $x_l$ as follows:
	\begin{equation}
	\begin{split}
	y_l = \omega(&L,P) \circ x_l + F (x_l, W_l)\\
	&x_{l+1} = f(y_l)
	\end{split}
	\label{eq:residual_formula2}
	\end{equation}
	in which $\circ$ denotes the Hadamard product symbol and the weight values gradually decrease when pixels $P$ get farther away from the facial landmark points $L$. This shows that having a more complex function than a simple shortcut (identity mapping) can help the network to extract more effective features in less number of layers which results in a shallower network and less number of parameters to be trained.
	
	In Equation~\eqref{eq:residual_formula3}, it can be seen that the identity bypass mapping ($x$) is a simple choice and is not contributing to feature learning. In fact, the original motivation for using $x$ in the residual connection was to have bounded feedbacks from the loss layer to every other layers of the network. Building on this observation, we studied developing more complex residual connections with bounded gradient which enrich feature learning through the residual parts of the network. This results in richer feature maps and therefore shallower networks. We investigated several functions and replaced the shortcut path in the network with those functions. There are few limitations on choosing the suitable function and not all the functions can be used, as the network will not converge otherwise. The reason behind this is that in the training phase, we need to calculate the gradient. An improper choice of the function will cause facing with either vanishing or exploding gradient. To have a better understanding of this concept we start with the ResNet's residual unit formulation. In this case, since we have an identical mapping of the inputs for the function $H(x_l)$, Equation~\eqref{eq:residual_formula} and its derivative will be re-written as follows:
	\begin{equation}
	\begin{split}
	y_l = x& + F (x_l, W_l)\\
	y_l' = 1& + F' (x_l, W_l)
	\end{split}
	\label{eq:residual_formula3}
	\end{equation}
	It is obvious that $H(x_l)=x$ is differentiable and its derivative is constant which means that it is also bounded. This allows the ResNet to converge and prevents the vanishing/exploding gradient problem. Therefore, any other function that is the replacement of $x$ needs to have the same properties. 
	
\begin{figure*}
		\centering
		\subfigure[$x-\log(e^x +1)$]{\label{fig:f1(x)}\includegraphics[width=0.23\linewidth]{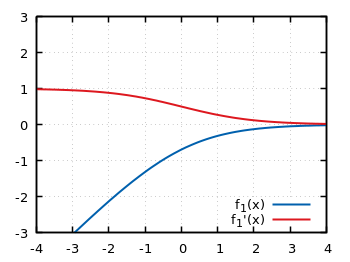}}
		\subfigure[$ x\tan^{-1}(x)-\frac{1}{2}\log(x^2+1)$]{\label{fig:f2(x)}\includegraphics[width=0.23\linewidth]{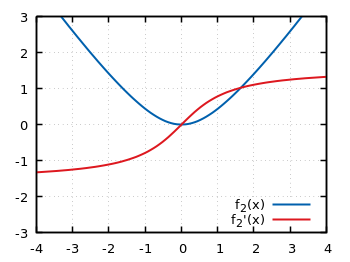}}
		\subfigure[$\tan^{-1}(x)$]{\label{fig:f3(x)}\includegraphics[width=0.23\linewidth]{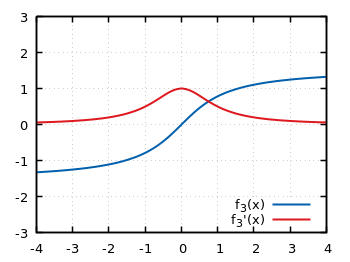}}
		\caption{Plots of proposed functions and their derivatives for the BReG-Net (best in color)}\label{fig:functions}
	\end{figure*}

	We observed several functions that have the aforementioned properties. Our experiments show that by incorporating any of these functions, the network will still converge and this is not surprising, based on the aforementioned argument. Hence, it is a matter of choosing the right function to have the best results for the facial expression task and valence/arousal prediction. Among the functions we investigated, the followings showed the most promising results:
	\begin{equation}
	\begin{split}
	H_1(x)&=x-\log(e^x +1), H_1'(x)=\frac{1}{1+e^x}\\
	H_2(x)&= x\tan^{-1}(x)-\frac{1}{2}\log(x^2+1), H_2'(x)= \tan^{-1}(x) \\
	H_3(x)&=\tan^{-1}(x), H_3'(x)=\frac{1}{1+x^2}
	\end{split}
	\label{eq:modification_funnctions}
	\end{equation} 
	and the corresponding derivative of these functions is as follows:

	Figure~\ref{fig:functions} shows the plots of these three functions and their derivatives. As shown, all of these functions are differentiable at any point and their derivatives are also bounded which shows that previously mentioned conditions hold for all of these functions. We call our network \textbf{B}ounded \textbf{Re}sidual \textbf{G}radient \textbf{Net}work (BReG-Net). Figure~\ref{fig:CResNet} shows the resulting building block of BReG-Net module. In our proposed network, similar to ResNet, we have dimension reductions of the tensor, achieved by down sampling (stride 2) on the first convolution layer of $F(x)$ (Figure~\ref{fig:CResNetDS}). As explained in the experiments section, we stack up 39 layers of these blocks in all of our experiments and compare the results on different databases.
	
	\begin{figure*}
		\centering
		\includegraphics[width=.58\textwidth]{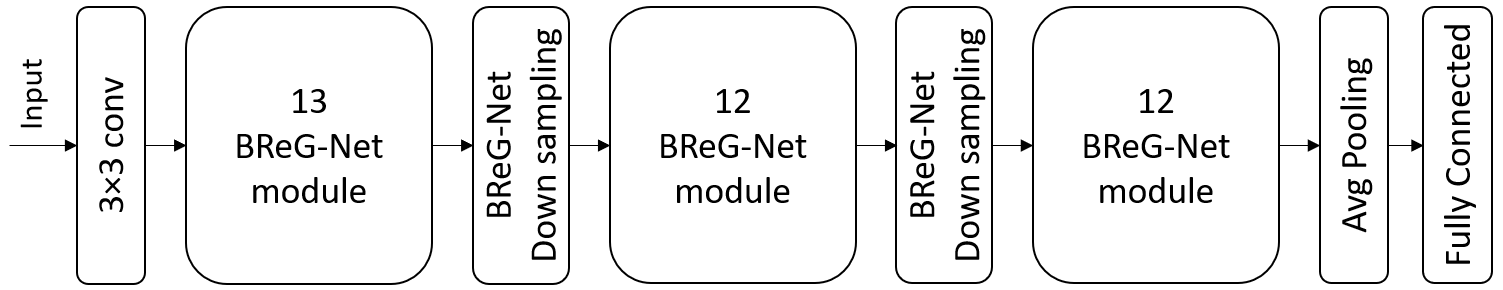}
		\caption{General architecture of the proposed method}
		\label{fig:archite}
	\end{figure*}
	
	\subsection{Weighted loss}
	
	Facial expression databases are usually highly skewed. This form of imbalance is commonly referred to as \emph{intrinsic} variation, i.e., it is a direct result of the nature of expressions in the real world. This phenomenon exists in both the categorical and dimensional models of affect. 
	For instance, in AffectNet database well represented categories like happiness have almost 30 times more number of samples than less represented categories like contempt. The problem of learning from imbalanced data has two downsides. First, training data with an imbalanced distribution often causes learning algorithms to perform poorly on the less represented category~\cite{he2009learning}. Second, imbalanced test/validation data can affect the performance metrics drastically resulting in an unrealistic image of method's performance. Jeni \textit{et al.}~\cite{jeni2013facing} studied the influence of skew on imbalanced datasets. This study shows that except for of area under the ROC curve (AUC), many other evaluation metrics such as accuracy, F1-score, Cohen’s kappa~\cite{cohen1960}, Krippendorf’s alpha ~\cite{krippendorff1970estimating}, and area under Precision-Recall curve (AUC-PR) are affected by skewed distributions dramatically. In order to minimize skew-biased estimates of performance, the study suggests reporting both skew-normalized metrics as well as the original evaluation.

	In the result section, we report the skew-normalized metrics of our methods in  addition to Matthews Correlation Coefficient (MCC)~\cite{matthews1975comparison} and Positive Predictive Value (PPV) which is often called precision. Moreover, in order to improve the recognition rate of different categories of emotions in our methods, we assign higher priority to the less represented categories of the databases in the loss calculation layer of our networks. We weigh the loss function for each of the classes by their relative proportion in the training dataset. In other words, the loss function highly penalizes the networks for misclassifying examples from under-represented categories, while it penalizes the networks less for misclassifying examples from well-represented categories. The entropy loss formulation for a training example $(x_i, l)$ is defined as:
	\begin{equation}
	\label{eq:infoGainLoss}
	E = - \sum_{i=1}^{K}{H_{l,i} log(\hat{p_i})}
	\end{equation}
	where $H_{l,i}$ denotes row $l$ penalization factor of class $i$. $K$ is the number of classes and $\hat{p_i}$ is the 
	predictive softmax with values in interval $[0,1]$ indicating the predicted probability of each class as:
	\begin{equation}
	\hat{p_i} = \frac{exp(z_i)}{\sum_{j=1}^{K}{exp(z_j)}}
	\end{equation}

	When $H = I$ ($I$ is the identity matrix), the proposed weighted-loss approach will turn to the traditional cross-entropy loss function. In other words, if the training data is completely balanced, the weighted-loss method is equal to the conventional cross-entropy loss function. We implemented this loss function in our TensorFlow model and we define the diagonal matrix $H_{ij}$ as:
	\begin{equation}
	\label{eq:H_Matrix}
	H_{ij} = 
	\begin{cases}
	\frac{f_{min}}{f_{i}},				& \text{if } i = j\\
	0,                   		& \text{otherwise}
	\end{cases}
	\end{equation}
	where $f_i$ is the number of samples in the $i^{{th}}$ category and $f_{min}$ is the number of samples in the least-represented category. As mentioned earlier, this will cause the loss function to highly penalize the network for misclassifying examples from under-represented categories. In the results section we show that this improves the network recognition of under represented categories and has an overall better recognition rate.  
	
	\section{EXPERIMENTS AND RESULTS}\label{sec:expresults}
	
	In this section, we briefly review the face databases used for evaluating our proposed method. We then report the results of our experiments using these databases evaluated on different metrics on both categorical and dimensional model of affect.
	
	\subsection{Face databases}
	As noted earlier, many of the traditional facial expression databases are assembled in a controlled environment while for developing a practical methods, these databases do not yield satisfying results. Therefore, we chose databases that are captured in the wild setting which contain a variety of backgrounds, lighting, pose, subject ethnicity, etc. These databases are AffectNet~\cite{mollahosseini2017affectnet}, Affect-in-Wild~\cite{zafeiriou2017aff}, and FER2013~\cite{FER2013} of which AffectNet contains labels of both categorical and dimensional models. Affect-in-Wild contains only labels of dimensional model, and FER2013 contains only labels of categorical model. AffectNet contains more than one million facial images collected from the Internet by querying three major search engines using 1250 emotion related keywords in six different languages. Affect-in-Wild contains 300 videos of different subjects watching videos of various TV shows and movies. FER2013 was created using the Google image search API. Faces  are  labeled with any of the six basic expressions, along with neutral. The resulting database contains 35,887 images in the wild settings.

	\subsection{Evaluation metrics of dimensional model}
	In order to evaluate our methods, we calculate and report Root Mean Square Error (RMSE), Correlation Coefficient (CC), Concordance Correlation Coefficient (CCC), and Sign AGReement (SAGR) metrics for our methods. In the following, we briefly describe the definitions of these metrics.
	
	Root Mean Square Error (RMSE) is the most common evaluation metric in a continuous domain which is
	defined as:
	\begin{equation}
	RMSE = \sqrt{\frac{1}{n} \sum_{i=1}^{n}(\hat{\theta}_i-\theta_i)^2}
	\end{equation}
	where $\hat{\theta}_i$ and $\theta_i$ are the prediction and the ground-truth of $i^{{th}}$ sample, and $n$ is the number of samples. RMSE-based evaluation metrics can heavily weigh the outliers~\cite{bermejo2001oriented}, and they do not consider covariance of the data. 
	
	Pearson's Correlation Coefficient (CC) overcomes this problem~\cite{nicolaou2011continuous} and it is defined as: 
	\begin{equation}
	CC = \frac{COV\{\hat{\theta}, \theta\}}{\sigma_{\hat{\theta}}\sigma_{\theta}} = 
	\frac{E [(\hat{\theta}-\mu_{\hat{\theta}})(\theta-\mu_{\theta})]}{\sigma_{\hat{\theta}}\sigma_{\theta}}
	\end{equation}
	where $COV$ is covariance function.
	
	Concordance Correlation Coefficient (CCC) is another metric~\cite{valstar2016avec} and combines CC with the square difference between the means of two compared time series:
	
	\begin{equation}
	\rho_c = \frac{2\rho \sigma_{\hat{\theta}} \sigma_{\theta}}{\sigma_{\hat{\theta}}^2 + \sigma_{\theta}^2 + (\mu_{\hat{\theta}} - \mu_\theta)^2}
	\end{equation}
	where $\rho$ is the Pearson correlation coefficient (CC) between two time-series (e.g., prediction and ground-truth), $\sigma_{\hat{\theta}}^2$ and $\sigma_{\theta}^2$ are the variance of each time series, $\sigma_{\hat{\theta}}$ and $\sigma_{\theta}$ are the standard deviation of each, and $\mu_{\hat{\theta}}$ and $\mu_{\theta}$ are the mean value of each. Unlike CC, the predictions that are well correlated with the ground-truth but shifted in value are penalized in proportion to the deviation in the CCC.

	The value of valence and arousal fall within the interval of [-1,+1] and correctly predicting their signs are essential in many emotion-prediction applications. Therefore, we use Sign AGReement (SAGR) metric as proposed in~\cite{nicolaou2011continuous} to evaluate the performance of a valence and arousal prediction system with respect to the sign agreement. SAGR is defined as:
	\begin{equation}
	SAGR = \frac{1}{n}\sum_{i=1}^{n}\delta{(sign(\hat{\theta}_i),sign(\theta_i))}
	\end{equation}
	where $\delta$ is the Kronecker delta function, defined as:
	\begin{equation}
	\delta{(a,b)} = 
	\begin{cases}
	1,				& a = b\\
	0,              & a \neq b
	\end{cases}
	\end{equation} 
	
	\subsection{Results}
\begin{figure}[tb]
		\centering
		\subfigure[Categorical]{\label{fig:depthcategorical}\includegraphics[width=.45\linewidth]{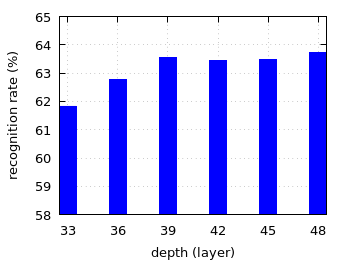}}
		\subfigure[Dimensional]{\label{fig:depthdimensional}\includegraphics[width=.45\linewidth]{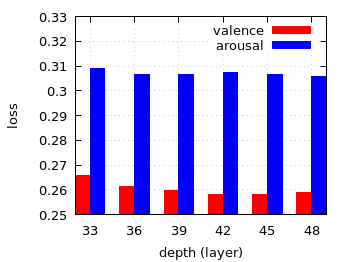}}
		\caption{Result of experimenting different depth on categorical and dimensional model (best in color)}\label{fig:depths}
	\end{figure}

	Figure~\ref{fig:archite} shows the general structure of the network. 
	Our experiments show that $H_3(x)$ yields better results in terms of both prediction rate and convergence speed. We also investigated a variety of BReG-Net architectures  with shallower and deeper depths. Our experiments indicated that when the network is too shallow, the number of parameters is not enough to distinguish the subtle facial muscle changes. Figure~\ref{fig:depths} shows the results of different depths in both categorical and dimensional models of affect while using $H_3(x)=\tan^{-1}(x)$ as residual function in our proposed method. Thus, we propose the architecture in Figure~\ref{fig:archite} for two tasks of prediction of categorical and dimensional model of affect. We provide the results of our experiment for each of these tasks separately. All of the proposed methods are implemented using a combination of TensorFlow~\cite{AABB01} and TfLearn~\cite{tflearn2016} toolboxes.  We used Momentum optimization method with a weight decay of 0.0001, and learning rate of 0.01. Mean square error is used for the loss function of the dimensional model experiments.
	
	\subsubsection{Categorical model}

	Table~\ref{Tab:resultsoffucntion} shows the results of our experiments with the three functions in Equation~\eqref{eq:modification_funnctions} as the residual function. We can see that $H_3(x)=\tan^{-1}(x)$ has the best result compared to the other functions. This was true throughout all of the experiments. Therefore, due to space limitation, all of the reported results from this point are the result of $H_3(x)$ function. Table~\ref{tab:categorical} shows the result of our experiments in the categorical model of affect on AffectNet and FER2013 databases. It can be seen that weighted loss further improves the recognition rates in both databases. However, weighted-loss is data dependent while our proposed method improves the recognition rate regardless of the distribution of the data. All of the reported numbers, are the result of our experiments only on the validation set of these databases as their test sets are not publicly available for any of the databases. As it can be seen, our proposed modification of the ResNet module achieves better recognition rates compared to ResNet-110 and it also outperforms the existing methods on both AffectNet and FER2013 databases. We need to mention that~\cite{mollahosseini2017affectnet} uses AlexNet, Wiles \emph{et al.}~\cite{wiles2018self} achieved 74.4 for AUC, and~\cite{c4} uses an Inception-based method to classify the expressions, and~\cite{tang2013deep} trained deep learning methods combined with SVMs. Our proposed method is considerably shallower than many of the methods proposed in the field. 
	
	In order to further investigate the effect of the weighted-loss method, we calculated F1-score, alpha, kappa, MCC, and PPV metrics in both cases of regular loss and weighted-loss. Tables~\ref{Tab:weighted_categorical} and~\ref{tab:notweighted_categorical} show the results for these losses, respectively.  The  skew  normalization  is  performed by  random  under-sampling  of  the  classes  in  the  test  set. This process is repeated 200 times, and the skew-normalized score  is  the  average  of  the  score  on  multiple  trials. It can be seen that in most cases there is an improvement of correlation in the weighted loss case which shows that our weighted loss addition to the network has a positive impact in recognition of different categories. It is important to note that the FER2013 database is an almost balanced database. Therefore, the reported results for original and skew-normalized cases have almost the same value.
	
	\begin{table*}[]
		\centering
		\caption{Recognition rate (\%) of proposed functions in Equation~\eqref{eq:modification_funnctions} in categorical model}
		\label{Tab:resultsoffucntion}
		\begin{tabular}{l|c|c|c|c|c|c|}
			\cline{2-7}
			& \multicolumn{2}{c|}{$H_1(x)$}                                                                                         & \multicolumn{2}{c|}{$H_2(x)$}                                                                                         & \multicolumn{2}{c|}{$H_3(x)$}                                                                                         \\ \cline{2-7} 
			& \begin{tabular}[c]{@{}c@{}}regular\\  loss\end{tabular} & \begin{tabular}[c]{@{}c@{}}weighted\\  loss\end{tabular} & \begin{tabular}[c]{@{}c@{}}regular\\  loss\end{tabular} & \begin{tabular}[c]{@{}c@{}}weighted\\  loss\end{tabular} & \begin{tabular}[c]{@{}c@{}}regular\\  loss\end{tabular} & \begin{tabular}[c]{@{}c@{}}weighted\\  loss\end{tabular} \\ \hline
			\multicolumn{1}{|l|}{\textbf{AffectNet}} & 57.37                                                   & 58.83                                                    & 59.43                                                   & 64.02                                                    & 60.03                                                   & 63.54                                                    \\ \hline
			\multicolumn{1}{|l|}{\textbf{FER2013}}   & 65.80                                                   & 66.21                                                    & 65.16                                                   & 67.66                                                    & 68.74                                                   & 69.49                                                    \\ \hline
		\end{tabular}
	\end{table*}
	
	\begin{table*}[]
		\centering
		\caption{Recognition rates (\%) in categorical model of affect}
		\label{tab:categorical}
		\begin{tabular}{l|c|c|c|c|}
			\cline{2-5}
			\multirow{2}{*}{}                        & \multirow{2}{*}{ResNet-110} & \multicolumn{2}{c|}{proposed method} & \multicolumn{1}{l|}{state-of-the-art} \\ \cline{3-4}
			&                                  & regular loss     & weighted loss     & methods                               \\ \hline
			\multicolumn{1}{|l|}{\textbf{AffectNet}} & 58.20                            & 60.03            & \textbf{63.54}    & 58.0~\cite{mollahosseini2017affectnet}, 57.31~\cite{zeng2018facial}                                     \\ \hline
			\multicolumn{1}{|l|}{\textbf{FER2013}}   & 66.48                            & 68.74            & \textbf{69.49}    & 69.3~\cite{tang2013deep},  66.4~\cite{c4}                                      \\ \hline
		\end{tabular}
	\end{table*}

	\begin{table*}[]
		\centering
		\caption{Results of  weighted-loss experiments on categorical model of affect}
		\label{Tab:weighted_categorical}
		\begin{tabular}{lclclclclclclclclclcl}
			\cline{2-21}
			\multicolumn{1}{l|}{\multirow{2}{*}{}}   & \multicolumn{4}{c|}{F1-score}                           & \multicolumn{4}{c|}{kappa}                            & \multicolumn{4}{c|}{alpha}                            & \multicolumn{4}{c|}{MCC}                              & \multicolumn{4}{c|}{PPV}                              \\ \cline{2-21} 
			\multicolumn{1}{l|}{}                    & \multicolumn{2}{c|}{Orig*} & \multicolumn{2}{c|}{Norm*} & \multicolumn{2}{c|}{Orig} & \multicolumn{2}{c|}{Norm} & \multicolumn{2}{c|}{Orig} & \multicolumn{2}{c|}{Norm} & \multicolumn{2}{c|}{Orig} & \multicolumn{2}{c|}{Norm} & \multicolumn{2}{c|}{Orig} & \multicolumn{2}{c|}{Norm} \\ \hline
			\multicolumn{1}{|l|}{\textbf{AffectNet}} & \multicolumn{2}{c|}{0.63}  & \multicolumn{2}{c|}{0.68}  & \multicolumn{2}{c|}{0.58} & \multicolumn{2}{c|}{0.63} & \multicolumn{2}{c|}{0.58} & \multicolumn{2}{c|}{0.64} & \multicolumn{2}{c|}{0.59} & \multicolumn{2}{c|}{0.64} & \multicolumn{2}{c|}{0.62} & \multicolumn{2}{c|}{0.71} \\ \hline
			\multicolumn{1}{|l|}{\textbf{FER2013}}   & \multicolumn{2}{c|}{0.67}  & \multicolumn{2}{c|}{0.67}  & \multicolumn{2}{c|}{0.62} & \multicolumn{2}{c|}{0.62} & \multicolumn{2}{c|}{0.62} & \multicolumn{2}{c|}{0.62} & \multicolumn{2}{c|}{0.62} & \multicolumn{2}{c|}{0.62} & \multicolumn{2}{c|}{0.69} & \multicolumn{2}{c|}{0.68} \\ \hline
			\multicolumn{21}{l}{*Orig and Norm stand for \textbf{Orig}inal and skew-\textbf{Norm}alized, respectively.}                           
		\end{tabular}
	\end{table*}

	\begin{table*}[]
		\centering
		\caption{Results of regular-loss experiments on categorical model of affect}
		\label{tab:notweighted_categorical}
		\begin{tabular}{lclclclclclclclclclcl}
			\cline{2-21}
			\multicolumn{1}{l|}{\multirow{2}{*}{}}   & \multicolumn{4}{c|}{F1-score}                           & \multicolumn{4}{c|}{kappa}                            & \multicolumn{4}{c|}{alpha}                            & \multicolumn{4}{c|}{MCC}                              & \multicolumn{4}{c|}{PPV}                              \\ \cline{2-21} 
			\multicolumn{1}{l|}{}                    & \multicolumn{2}{c|}{Orig*} & \multicolumn{2}{c|}{Norm*} & \multicolumn{2}{c|}{Orig} & \multicolumn{2}{c|}{Norm} & \multicolumn{2}{c|}{Orig} & \multicolumn{2}{c|}{Norm} & \multicolumn{2}{c|}{Orig} & \multicolumn{2}{c|}{Norm} & \multicolumn{2}{c|}{Orig} & \multicolumn{2}{c|}{Norm} \\ \hline
			\multicolumn{1}{|l|}{\textbf{AffectNet}} & \multicolumn{2}{c|}{0.58}  & \multicolumn{2}{c|}{0.60}  & \multicolumn{2}{c|}{0.52} & \multicolumn{2}{c|}{0.54} & \multicolumn{2}{c|}{0.52} & \multicolumn{2}{c|}{0.54} & \multicolumn{2}{c|}{0.52} & \multicolumn{2}{c|}{0.54} & \multicolumn{2}{c|}{0.58} & \multicolumn{2}{c|}{0.62} \\ \hline
			\multicolumn{1}{|l|}{\textbf{FER2013}}   & \multicolumn{2}{c|}{0.67}  & \multicolumn{2}{c|}{0.68}  & \multicolumn{2}{c|}{0.61} & \multicolumn{2}{c|}{0.62} & \multicolumn{2}{c|}{0.61} & \multicolumn{2}{c|}{0.62} & \multicolumn{2}{c|}{0.62} & \multicolumn{2}{c|}{0.62} & \multicolumn{2}{c|}{0.69} & \multicolumn{2}{c|}{0.69} \\ \hline
			\multicolumn{21}{l}{*Orig and Norm stand for \textbf{Orig}inal and skew-\textbf{Norm}alized, respectively.}          
		\end{tabular}
	\end{table*}
	
	\subsubsection{Dimensional model}
\begin{figure}
		\centering
		\subfigure[AffectNet]{\label{fig:loss1}\includegraphics[width=.45\linewidth]{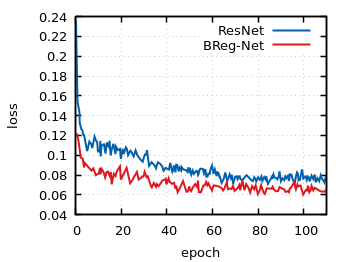}}
		\subfigure[Affect-in-Wild]{\label{fig:loss3}\includegraphics[width=.45\linewidth]{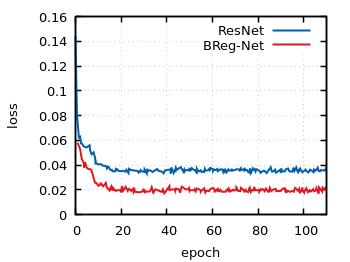}}\\
		\subfigure[AffectNet]{\label{fig:affectnet}\includegraphics[width=.45\linewidth]{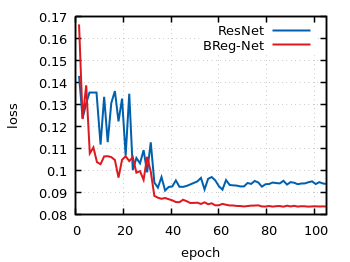}}
		\subfigure[Affect-in-Wild]{\label{fig:affectinwild}\includegraphics[width=.45\linewidth]{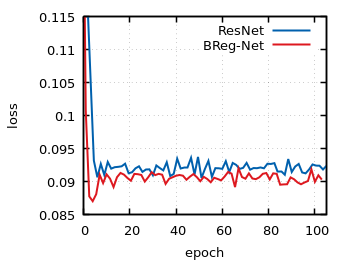}}
		
		\caption{Mean square loss of \textbf{training} (a and b) and \textbf{validation} (c and d) for ResNet-110 and BReG-Net (best in color)}\label{fig:loss}
	\end{figure}

	Table~\ref{tab:dimentional} shows the results of our experiments in the dimensional model of affect on the validation set of the AffectNet and Affect-in-Wild databases (test set was not released for either of the databases). It is important to point out that~\cite{mollahosseini2017affectnet} uses AlexNet, and~\cite{hasani2017facial_dimensional} uses an Inception-ResNet-based method to classify the expressions. The reported results are RMSE values, as other methods have only provided this metric in their work. Table~\ref{tab:dimentional} shows that our proposed method outperforms the state-of-the-art methods in terms of RMSE for both databases. Our results show significant improvement compared to methods reported in the AffectNet paper~\cite{mollahosseini2017affectnet}. Also, as shown in the categorical model experiments, we can see significant improvement using the BReG-Net comparing to ResNet-110. Figure~\ref{fig:loss} shows that our proposed method has a higher reduction rate compared to ResNet-110 and eventually reaches a lower loss value on both training and validation sets during training.
	
	In order to further investigate the effect of BReG-Net in the dimensional model of affect, we report the results by using the metrics of CC, CCC, and SAGR. Tables~\ref{Tab:weighted} and~\ref{Tab:notweighted} show the values of these metrics on BReG-Net and ResNet-110, respectively. It can be seen that the sign agreement is significantly improved when using BReG-Net, and also correlation of the predicted values is higher than the ones for ResNet. Also, we can see that predicted valence values have lower RMSE while have higher correlation with ground-truth compared to their corresponding arousal values. This is not surprising as RMSE and correlation coefficient measure two different aspects of distribution of the data. These tables also show that the Affect-in-Wild database is a more challenging database as the predicted values have less correlation with the ground-truth ones.
	
	In order to compare the computational cost of BReG-Net and ResNet, we recorded the computation time of training the model for one epoch on AffectNet database in categorical model. The average processing time of an epoch on AffectNet for BReG-Net with 4.9M parameters is 750.21 seconds and for ResNet-110 with 7.2M parameters is 836.04 seconds on a GeForce GTX 1080 Ti GPU. Therefore, our proposed method is trained considerably faster than ResNet-110 as it has less number of parameters to train.
	
	\begin{table*}[]
		\centering
		\caption{RMSE values of experiments on dimensional model of affect}
		\label{tab:dimentional}
		\begin{tabular}{l|c|c|c|c|c|c|}
			\cline{2-7}
			\multirow{2}{*}{}                             & \multicolumn{2}{c|}{ResNet-110} & \multicolumn{2}{c|}{proposed method} & \multicolumn{2}{c|}{\begin{tabular}[c]{@{}c@{}}state-of-the-art\\ methods\end{tabular}} \\ \cline{2-7} 
			& valence           & arousal          & valence          & arousal         & valence                          & \multicolumn{1}{l|}{arousal}                         \\ \hline
			\multicolumn{1}{|l|}{\textbf{AffectNet}}      & 0.2693            & 0.3082           & \textbf{0.2597}  & \textbf{0.3067} & 0.37~\cite{mollahosseini2017affectnet}                              & 0.41~\cite{mollahosseini2017affectnet}                                                  \\ \hline
			\multicolumn{1}{|l|}{\textbf{Affect-in-Wild}} & 0.2733            & 0.3309           & \textbf{0.2661}  & \textbf{0.3265} & 0.27~\cite{hasani2017facial_dimensional}                            & 0.36~\cite{hasani2017facial_dimensional}                                                 \\ \hline
		\end{tabular}
	\end{table*}

	\begin{table*}[]
		\centering
		\caption{Results of BReG-Net on dimensional model}
		\label{Tab:weighted}
		\begin{tabular}{l|c|c|c|c|c|c|}
			\cline{2-7}
			\multirow{2}{*}{}                             & \multicolumn{2}{c|}{CC} & \multicolumn{2}{c|}{CCC} & \multicolumn{2}{c|}{SAGR} \\ \cline{2-7} 
			& valence    & arousal    & valence     & arousal    & valence     & arousal     \\ \hline
			\multicolumn{1}{|l|}{\textbf{AffectNet}}      & 0.66       & 0.84       & 0.66        & 0.82       & 0.73        & 0.84        \\ \hline
			\multicolumn{1}{|l|}{\textbf{Affect-in-Wild}} & 0.45       & 0.40       & 0.43        & 0.34       & 0.63        & 0.77        \\ \hline
		\end{tabular}
	\end{table*}
	
	\begin{table*}[]
		\centering
		\caption{Results of ResNet-110 on dimensional model}
		\label{Tab:notweighted}
		\begin{tabular}{l|c|c|c|c|c|c|}
			\cline{2-7}
			\multirow{2}{*}{}                             & \multicolumn{2}{c|}{CC} & \multicolumn{2}{c|}{CCC} & \multicolumn{2}{c|}{SAGR} \\ \cline{2-7} 
			& valence    & arousal    & valence     & arousal    & valence     & arousal     \\ \hline
			\multicolumn{1}{|l|}{\textbf{AffectNet}}      & 0.66       & 0.84       & 0.63        & 0.82       & 0.66        & 0.84        \\ \hline
			\multicolumn{1}{|l|}{\textbf{Affect-in-Wild}} & 0.41       & 0.41       & 0.38        & 0.35       & 0.61        & 0.75        \\ \hline
		\end{tabular}
	\end{table*}
	
	\section{CONCLUSION}\label{sec:conclusion}
	This paper introduces BReG-Net, a new residual-based network architecture using a  differentiable and bounded gradient function instead of a shortcut path between the input and the output of the residual block for the task of affect estimation in both categorical and dimensional models of affect. Our experiments showed that recruiting more complex units will result in shallower networks with better performance. We also used weighted loss function in the categorical model, where our method gives higher priority to the under represented categories, resulting in a better recognition rate. We evaluated our  proposed  method  on three databases of facial images captured in wild settings. Our experiments showed that the proposed method outperforms state-of-the-art methods in both tasks.
	\section{Acknowledgement}
	This paper is based upon work partially supported by the National Science Foundation under Grant No. CNS-1427872. We also thank NVIDIA for donation of a GPU to the University of Denver.

	\bibliographystyle{ieee}
	\bibliography{egbib}

\begin{thebibliography}{10}\itemsep=-1pt

\bibitem{FER2013}
Challenges in representation learning: Facial expression recognition challenge.
\newblock
  http://www.kaggle.com/c/challenges-in-representation-learning-facial-expression-recognition-challenge.

\bibitem{bermejo2001oriented}
S.~Bermejo and J.~Cabestany.
\newblock Oriented principal component analysis for large margin classifiers.
\newblock {\em Neural Networks}, 14(10):1447--1461, 2001.

\bibitem{cohen1960}
J.~Cohen.
\newblock A coefficient of agreement for nominal scales.
\newblock {\em Educational and Psychological Measurement}, 20(1):37, 1960.

\bibitem{tflearn2016}
A.~Damien et~al.
\newblock Tflearn.
\newblock https://github.com/tflearn/tflearn, 2016.

\bibitem{c5}
P.~Ekman and W.~V. Friesen.
\newblock Constants across cultures in the face and emotion.
\newblock {\em Journal of personality and social psychology}, 17(2):124, 1971.

\bibitem{hasani2017facial_dimensional}
B.~Hasani and M.~H. Mahoor.
\newblock Facial affect estimation in the wild using deep residual and
  convolutional networks.
\newblock In {\em CVPR Workshops}, pages 1955--1962. IEEE, 2017.

\bibitem{hasani2017facial}
B.~Hasani and M.~H. Mahoor.
\newblock Facial expression recognition using enhanced deep 3d convolutional
  neural networks.
\newblock In {\em CVPR Workshops}, pages 2278--2288. IEEE, 2017.

\bibitem{he2009learning}
H.~He and E.~A. Garcia.
\newblock Learning from imbalanced data.
\newblock {\em IEEE Transactions on Knowledge \& Data Engineering},
  21(9):1263--1284, 2009.

\bibitem{he2015deep}
K.~He, X.~Zhang, S.~Ren, and J.~Sun.
\newblock Deep residual learning for image recognition.
\newblock {\em arXiv preprint arXiv:1512.03385}, 2015.

\bibitem{c6}
K.~He, X.~Zhang, S.~Ren, and J.~Sun.
\newblock Deep residual learning for image recognition.
\newblock In {\em CVPR}, June 2016.

\bibitem{he2016identity}
K.~He, X.~Zhang, S.~Ren, and J.~Sun.
\newblock Identity mappings in deep residual networks.
\newblock {\em arXiv preprint arXiv:1603.05027}, 2016.

\bibitem{he2015multimodal}
L.~He, D.~Jiang, L.~Yang, E.~Pei, P.~Wu, and H.~Sahli.
\newblock Multimodal affective dimension prediction using deep bidirectional
  long short-term memory recurrent neural networks.
\newblock In {\em Workshop on Audio/Visual Emotion Challenge}, pages 73--80.
  ACM, 2015.

\bibitem{c20}
S.~Ioffe and C.~Szegedy.
\newblock Batch normalization: Accelerating deep network training by reducing
  internal covariate shift.
\newblock {\em arXiv preprint arXiv:1502.03167}, 2015.

\bibitem{jeni2013facing}
L.~A. Jeni, J.~F. Cohn, and F.~De~La~Torre.
\newblock Facing imbalanced data--recommendations for the use of performance
  metrics.
\newblock In {\em ACII}, pages 245--251. IEEE, 2013.

\bibitem{koelstra2012deap}
S.~Koelstra, C.~Muhl, M.~Soleymani, J.-S. Lee, A.~Yazdani, T.~Ebrahimi, T.~Pun,
  A.~Nijholt, and I.~Patras.
\newblock Deap: A database for emotion analysis; using physiological signals.
\newblock {\em IEEE Transactions on Affective Computing}, 3(1):18--31, 2012.

\bibitem{krippendorff1970estimating}
K.~Krippendorff.
\newblock Estimating the reliability, systematic error and random error of
  interval data.
\newblock {\em Educational and Psychological Measurement}, 30(1):61--70, 1970.

\bibitem{AABB01}
A.~A. M.~Abadi et~al.
\newblock Tensorflow: Large-scale machine learning on heterogeneous.
\newblock {\em Software available from tensorflow. org, 1, 6}, 2015.

\bibitem{matthews1975comparison}
B.~W. Matthews.
\newblock Comparison of the predicted and observed secondary structure of t4
  phage lysozyme.
\newblock {\em Biochimica et Biophysica Acta (BBA)-Protein Structure},
  405(2):442--451, 1975.

\bibitem{c4}
A.~Mollahosseini, D.~Chan, and M.~H. Mahoor.
\newblock Going deeper in facial expression recognition using deep neural
  networks.
\newblock In {\em WACV}, pages 1--10. IEEE, 2016.

\bibitem{mollahosseini2017affectnet}
A.~Mollahosseini, B.~Hasani, and M.~H. Mahoor.
\newblock Affectnet: A database for facial expression, valence, and arousal
  computing in the wild.
\newblock {\em IEEE Transactions on Affective Computing}, 2017.

\bibitem{c53}
A.~Mollahosseini, B.~Hasani, M.~J. Salvador, H.~Abdollahi, D.~Chan, and M.~H.
  Mahoor.
\newblock Facial expression recognition from world wild web.
\newblock In {\em CVPR Workshops}, June 2016.

\bibitem{nicolaou2011continuous}
M.~A. Nicolaou, H.~Gunes, and M.~Pantic.
\newblock Continuous prediction of spontaneous affect from multiple cues and
  modalities in valence-arousal space.
\newblock {\em IEEE Transactions on Affective Computing}, 2(2):92--105, 2011.

\bibitem{rumelhart1986parallel}
D.~E. Rumelhart, J.~L. McClelland, P.~R. Group, et~al.
\newblock Parallel distributed processing: Explorations in the microstructures
  of cognition. volume 1: Foundations, 1986.

\bibitem{russell1980circumplex}
J.~A. Russell.
\newblock A circumplex model of affect.
\newblock {\em Journal of Personality and Social Psychology}, 39(6):1161--1178,
  1980.

\bibitem{c2}
C.~Shan, S.~Gong, and P.~W. McOwan.
\newblock Facial expression recognition based on local binary patterns: A
  comprehensive study.
\newblock {\em Image and Vision Computing}, 27(6):803--816, 2009.

\bibitem{c23}
C.~Szegedy, S.~Ioffe, and V.~Vanhoucke.
\newblock Inception-v4, inception-resnet and the impact of residual connections
  on learning.
\newblock {\em arXiv preprint arXiv:1602.07261}, 2016.

\bibitem{c18}
C.~Szegedy, W.~Liu, Y.~Jia, P.~Sermanet, S.~Reed, D.~Anguelov, D.~Erhan,
  V.~Vanhoucke, and A.~Rabinovich.
\newblock Going deeper with convolutions.
\newblock In {\em CVPR}, pages 1--9, 2015.

\bibitem{tang2013deep}
Y.~Tang.
\newblock Deep learning using linear support vector machines.
\newblock {\em arXiv preprint arXiv:1306.0239}, 2013.

\bibitem{tao2005affective}
J.~Tao and T.~Tan.
\newblock Affective computing: A review.
\newblock In {\em ACII}, pages 981--995. Springer, 2005.

\bibitem{valstar2016avec}
M.~Valstar, J.~Gratch, B.~Schuller, F.~Ringeval, D.~Lalanne, M.~T. Torres,
  S.~Scherer, G.~Stratou, R.~Cowie, and M.~Pantic.
\newblock Avec 2016-depression, mood, and emotion recognition workshop and
  challenge.
\newblock {\em arXiv preprint arXiv:1605.01600}, 2016.

\bibitem{wiles2018self}
O.~Wiles, A.~Koepke, and A.~Zisserman.
\newblock Self-supervised learning of a facial attribute embedding from video.
\newblock {\em arXiv preprint arXiv:1808.06882}, 2018.

\bibitem{zafeiriou2017aff}
S.~Zafeiriou, D.~Kollias, M.~A. Nicolaou, A.~Papaioannou, G.~Zhao, and
  I.~Kotsia.
\newblock Aff-wild: Valence and arousal in-the-wild challenge.
\newblock In {\em CVPR Workshop}, 2017.

\bibitem{zeng2018facial}
J.~Zeng, S.~Shan, and X.~Chen.
\newblock Facial expression recognition with inconsistently annotated datasets.
\newblock In {\em ECCV}, pages 222--237, 2018.

\end{thebibliography}
	
\end{document}